%% file: acl_latex.tex
\pdfoutput=1

\documentclass[11pt]{article}

\usepackage{acl}
\usepackage{times}
\usepackage{latexsym}

\usepackage[T1]{fontenc}

\usepackage[utf8]{inputenc}

\usepackage{microtype}

\usepackage{inconsolata}

\usepackage{graphicx}

\usepackage{tabularx} 

\usepackage{inconsolata}

\usepackage{graphicx}
\usepackage{multirow}
\usepackage{colortbl}
\usepackage{amsmath}
\usepackage{booktabs}
\usepackage{amssymb}
\usepackage{pifont}
\usepackage{enumitem}
\usepackage{todonotes}
\usepackage[most]{tcolorbox}
\usepackage{listings}
\usepackage{floatpag} 
\usepackage{multicol}
\usepackage{lipsum}
\usepackage{hyperref}
\usepackage{xcolor}

\usepackage{longtable}
\usepackage{tcolorbox}
\definecolor{Highlight}{HTML}{39b54a}
\newcommand{\hl}[1]{\textcolor{Highlight}{#1}}
\newcommand{\hlr}[1]{\textcolor{red}{#1}}
\title{\includegraphics[width=0.04\textwidth]{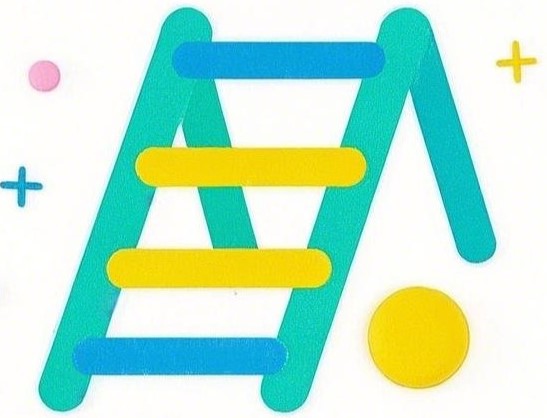} \textcolor{blue}{M}\textcolor{orange}{P}\textcolor{purple}{Bench}: A Comprehensive \textcolor{blue}{Multimodal} Reasoning \textcolor{purple}{Benchmark} for \textcolor{orange}{Process} Errors Identification}

\author{Zhaopan Xu$^{1,2}$, ~Pengfei Zhou$^{3}$,~Jiaxin Ai$^{4}$, ~Wangbo Zhao$^{3}$, ~Kai Wang$^{3}$,~Xiaojiang Peng$^{5}$\\ \textbf{~Wenqi Shao$^{2,4}$,~Hongxun Yao$^{1}$\thanks{~Corresponding author.},~Kaipeng Zhang$^{2,4}$}\footnotemark[1]\\
   $^{1}$HIT, 
   $^{2}$Shanghai AI Laboratory,$^{3}$NUS,$^{4}$Shanghai Innovation Institude,$^{5}$SZTU \\
  \texttt{h.yao@hit.edu.cn, zhangkaipeng@pjlab.org.cn}\\
\url{https://mpbench.github.io}}

\begin{document}
\maketitle

\input{text/0_abstract}

\input{text/1_introduction}

\input{text/2_relatedwork}

\input{text/3_method}

\input{text/4_experiment}

\input{text/5_conclusion}

\section*{Limitations}

Despite our best efforts throughout the entire benchmark construction process, MPBench may still contain inaccurate labels of error locations, particularly for the more challenging complex math problems.

\bibliography{custom}

\clearpage
\appendix

\newpage

\input{text/6_appendix}

\end{document}

%% file: text/0_abstract.tex
\begin{abstract}

Reasoning is an essential capacity for large language models (LLMs) to address complex tasks, where the identification of process errors is vital for improving this ability. Recently, process-level reward models (PRMs) were proposed to provide step-wise rewards that facilitate reinforcement learning and data production during training and guide LLMs toward correct steps during inference, thereby improving reasoning accuracy. However, existing benchmarks of PRMs are text-based and focus on error detection, neglecting other scenarios like reasoning search. To address this gap, we introduce MPBench, a comprehensive, multi-task, multimodal benchmark designed to systematically assess the effectiveness of PRMs in diverse scenarios. MPBench employs three evaluation paradigms, each targeting a specific role of PRMs in the reasoning process: (1) Step Correctness, which assesses the correctness of each intermediate reasoning step; (2) Answer Aggregation, which aggregates multiple solutions and selects the best one; and (3) Reasoning Process Search, which guides the search for optimal reasoning steps during inference. Through these paradigms, MPBench makes comprehensive evaluations and provides insights into the development of multimodal PRMs.

\end{abstract}

%% file: text/1_introduction.tex
\section{Introduction}
\label{submission}
Artificial intelligence (AI) has not yet effectively addressed complex reasoning tasks such as mathematics, programming, and planning. A barrier to achieving artificial general intelligence (AGI) remains this. To address this challenge, the recent release of the GPT-o1~\cite{openai_o1_2024} proposed learning to reason to master human-like reasoning processes. Unlike traditional LLMs' thinking, reasoning models generate long chains of thought. Consequently, recent research has increasingly paid attention to process-level (i.e., stepwise) reasoning analysis and further introduced various reasoning models to achieve significant enhancements in tasks such as mathematics and code generation.

Identifying process errors can facilitate effective reasoning trajectories. PRMs, introduced by~\cite{uesato2022solving} and~\cite{lightman2023let}, can provide stepwise feedback for multi-step reasoning results. By rewarding intermediate steps, PRMs can enhance LLMs' reasoning capabilities during both training and inference. Specifically, PRMs facilitate reinforcement learning or chain-of-thought data generation by offering fine-grained rewards, thus reducing the dependence on human-annotated data and improving performance.  Moreover, during inference, PRMs guide LLMs in evaluating and exploring intermediate ``thoughts," promoting the generation of more deliberate reasoning steps and ultimately leading to improved reasoning accuracy.

\input{tables/prmbench}

Despite the critical role of PRMs, they have not been adequately evaluated. Best-of-N performance, a widely employed evaluation paradigm, is time-consuming, lacks finer-grained inspection, and its evaluation reliability can be significantly affected by the underlying solution generation model. This limited scope of evaluation hinders a comprehensive understanding of PRMs' potential to enhance complex reasoning tasks. Recent benchmarks~\cite{qwenprocessbench,song2025prmbench} were introduced to evaluate PRMs. However, they typically focus on a single scenario, such as step-by-step error identification, evaluating the ability of PRMs to detect erroneous steps accurately. These evaluation paradigms inadequately assess PRM performance in some scenarios. For example, during the reasoning process search in LLM inference, PRMs are tasked with selecting the correct step from candidates without access to the complete reasoning trajectory and the final answer. Furthermore, whereas current PRM benchmarks are text-centric, multimodal benchmarks are deficient, while multimodal contents are common in real-world tasks.

These issues motivate our development of \textbf{MPBench}, a comprehensive multimodal benchmark for evaluating the efficacy of multimodal PRMs across three scenarios. MPBench focuses on three key evaluation paradigms: \textbf{Step Correctness}, which evaluates the ability of PRMs to evaluate the correctness of each intermediate reasoning step and provide stepwise rewards to support reinforcement learning; \textbf{Answer Aggregation}, where PRMs aggregate per-step scores from multiple candidate solutions to select the best one; and \textbf{Reasoning Process Search}, which examines the ability of PRMs to guide the search for optimal reasoning steps by enabling a structured exploration of potential solutions. With 9,745 fine-grained data instances across six sub-categories, MPBench offers a robust framework for comprehensively assessing PRM performance in real-world reasoning tasks, providing valuable insights into their role in improving the reasoning capabilities of MLLMs.

We conduct extensive experiments on MPBench with 12 MLLMs (prompted as critic models), including closed-source models, GPT-4o and Gemini-2.0, and open-source models like InternVL~\cite{chen2024expanding}, QWenVL~\cite{qwen2.5-VL}, and QVQ~\cite{qvq-72b-preview}. They struggle to achieve satisfactory results, and we believe our benchmarks benefit the future development of multimodal PRMs and process-level analysis. Our key contributions are summarized as follows:

\begin{itemize}
    \item We present MPBench, the first comprehensive multimodal process-level reward model benchmark, comprising 9,745 fine-grained data instances across diverse subjects, tasks, and challenges.

    \item MPBench incorporates three distinct evaluation paradigms that comprehensively assess the role of PRMs in enhancing MLLM reasoning during both training and inference. These paradigms include step correctness, answer aggregation, and reasoning process search.

    \item We in-depth analyze the performance of 12 MLLMs and reveal distinct performance characteristics across different scenarios, providing valuable insights to assist future research on the development of multimodal PRMs.
\end{itemize}

%% file: tables/prmbench.tex
\newcommand{\tabincell}[2]{\begin{tabular}{@{}#1@{}}#2\end{tabular}}
\begin{table*}[t]
  \centering
  \vspace{1mm}

  \resizebox{\textwidth}{!}{
    \begin{tabular}{lcccccc}
    \toprule
     & \tabincell{c}{\textbf{PRM}\\ \textbf{Benchmarks?} } & \tabincell{c}{\textbf{Multimodal}\\ \textbf{Benchmarks?}}&  \tabincell{c}{\textbf{Evaluation} \\ \textbf{Paradigms}} & \tabincell{c}{\textbf{Step} \\ \textbf{Annotation}}   & \textbf{Annotator} & \tabincell{c}{\textbf{Test Case}\\ \textbf{Size}}  \\
    \midrule
    MR-GSM8K \cite{mr_gsm8k}& \ding{55}& \ding{55}& 1 &\ding{51} &  Human & 2,999  \\
    CriticBench \cite{lin2024criticbench}& \ding{55}& \ding{55} & 1 & \ding{55}  & - & - \\
    MathCheck-GSM \cite{MathCheck_GSM}& \ding{55}  & \ding{55} & 1 & \ding{51}  & Synthetic & 516 \\
    M$^3$CoT \cite{chen2024m} & \ding{55}& \ding{55}& 1 &\ding{51} & Human & 5,975  \\
    ProcessBench \cite{qwenprocessbench}& \ding{51} & \ding{55} & 1 & \ding{51} & Human & 3,400  \\

    \textsc{PRMBench~\cite{song2025prmbench}}& \ding{51}& \ding{55}  & 1 &\ding{51} & Synthetic + Human & 6,216  \\ 
    \midrule
    \textsc{MPBench (Ours)}& \ding{51}& \ding{51}  & 3 &\ding{51} & Synthetic + Human & 9,745  \\ 
    
    \bottomrule
    \end{tabular}
  }
    \caption{Comparison between reasoning-related LLM benchmarks with our MPBench benchmark. 
  }
  \label{tab:comparison}
\end{table*}

%% file: text/2_relatedwork.tex
\section{Related Work}

\subsection{Reasoning Benchmarks}

GSM8K~\cite{cobbe2021training} and MATH~\cite{hendrycks2021measuring} have served as prominent benchmarks for evaluating the mathematical reasoning capabilities of LLMs, primarily focusing on assessing the final correctness of generated solutions. Subsequent work has explored synthesizing solutions and evaluating intermediate steps. For instance, MathCheck~\cite{MathCheck_GSM} focuses on judging the correctness of individual reasoning steps.  CriticBench~\cite{lin2024criticbench} evaluates language models' abilities to critique solutions and extends evaluation to various reasoning tasks. With the growing interest in the development of process reward models (PRMs) to enhance model reasoning capabilities, several benchmarks are specifically designed for PRMs. such as ProcessBench~\cite{qwenprocessbench} and PRMBench~\cite{song2025prmbench}.

These reasoning benchmarks are primarily text-based.  With the increasing demand for evaluating multimodal reasoning, there has been a drive to develop multimodal reasoning benchmarks across diverse domains~\cite{zeng2024mrben}. However, research on benchmarks specifically tailored to multimodal PRMs remains limited. To address this gap, we introduce MPBench to benchmark multimodal process reward models. As highlighted in Table~\ref{tab:comparison}, MPBench encompasses a wide range of tasks, along with a large-scale, diverse set of erroneous steps, enabling a more comprehensive evaluation of multimodal PRMs' reasoning capabilities.

\begin{figure*}[!t]
\centering
  \resizebox{1\linewidth}{!} {
    \includegraphics{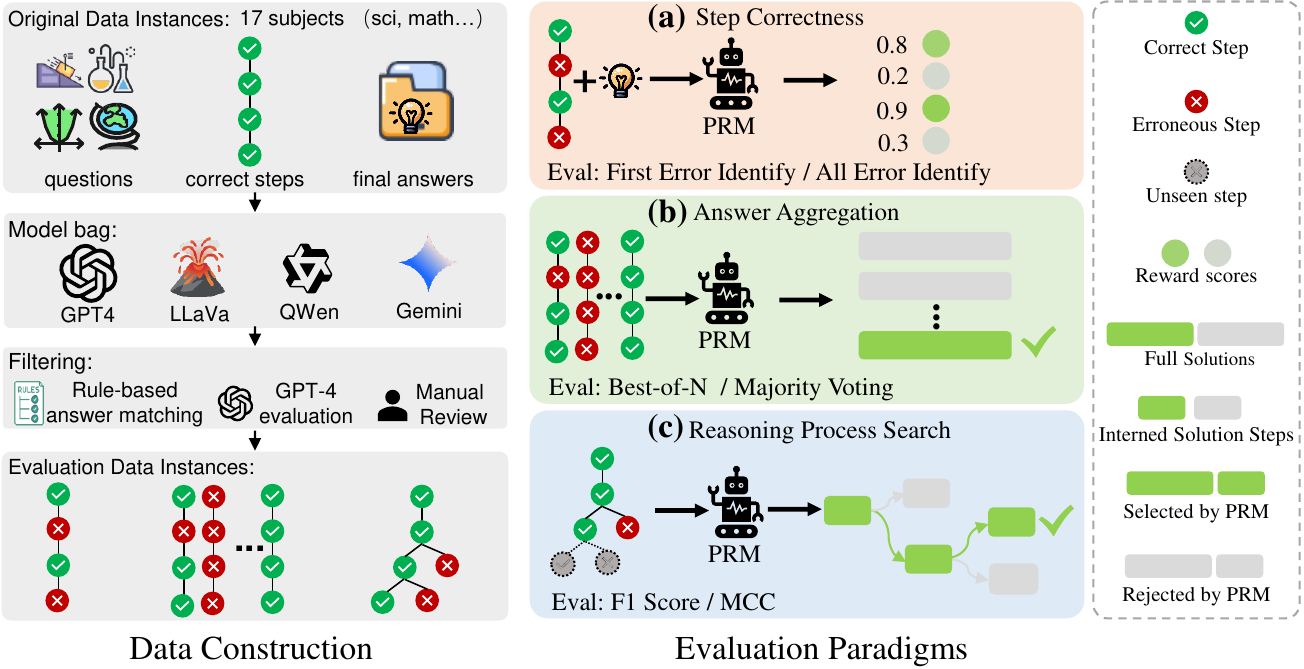}
  }
  \caption{An overview of our MPBench. Left: data curation procedure. Right: evaluation paradigms: Step Correctness, Answer Aggregation, and Reasoning Process Search, highlighting the assessment of PRM performance through various tasks such as identifying errors, aggregating answers, and guiding reasoning steps}
    \label{fig:main}
  \vspace{-4mm}
\end{figure*}
\subsection{Process Reward Models}
The increasing prevalence of Process Reward Models (PRMs) highlights their crucial role in enhancing reasoning by providing step-wise rewards that facilitate reinforcement learning and data generation. To evaluate the accuracy of these step-wise rewards, benchmarks like ProcessBench and PRMBench have been developed. These benchmarks focus on constructing reasoning processes with erroneous steps for mathematical problems and evaluating the PRMs' ability to identify the first error or the fine-grained error category of each step.
Recently, GPT-o1 has shown impressive reasoning trajectories, highlighting the potential of scaling test-time computation to enhance LLM reasoning accuracy. Building on this idea, several studies~\cite{snell2024scaling,xiang2024atomthink} have explored using PRMs to guide search within the interactive process during inference, further boosting model reasoning capabilities. However, evaluating PRMs in this application scenario remains limited. This is primarily due to the need for group and tree-structured reasoning process data, which existing PRM benchmarks do not adequately address. This gap in evaluation is precisely one of the aspects our work intends to address.

%% file: text/3_method.tex
\section{MPBench}
\label{sec:MPBench}

MPBench employs three evaluation paradigms, each targeting a specific role of PRMs in enhancing reasoning:

\paragraph{Step Correctness:} This paradigm evaluates the PRMs' ability to judge the correctness of each intermediate step within a solution, as illustrated in Fig.~\ref{fig:main}a.  Accurate step correctness evaluation enables PRMs to provide step-wise rewards, facilitating reinforcement learning for LLM reasoning~\cite{reasoneval} and reducing reliance on human-annotated data.

\paragraph{Answer Aggregation:} Given multiple solutions from the generator, PRMs are tasked with aggregating the per-step scores for each answer to determine the best candidate solution (Fig~\ref{fig:main} b). This capability allows PRMs to assess the overall quality of each solution and guide the model in selecting the most appropriate response from a set of candidates.

\paragraph{Reasoning Process Search:} This paradigm evaluates the PRMs' ability to guide the search for optimal reasoning steps during inference. As illustrated in Fig.~\ref{fig:main}c, PRMs provide step-wise predictions that enable a tree search over the solution space. This guided search encourages LLMs to generate more deliberate reasoning steps during inference, ultimately leading to improved reasoning accuracy.

MPBench includes 9,745 fine-grained instances across six sub-categories, designed to comprehensively evaluate models across the three key evaluation paradigms, which enables a robust assessment of PRM in handling realistic reasoning.

\subsection{Evaluation Objectives}
\subsubsection{Step Correctness}
\label{sec:step_correctness}

This assessment takes annotated question-solution pairs and corresponding ground truth answers as input. The task is to evaluate the correctness of each individual step, generating step-level scores.  A predefined threshold is then applied to these scores, yielding a binary prediction of step correctness. Consequently, this evaluation is framed as a binary classification problem, with the F1 score serving as the evaluation metric. Following ProcessBench~\cite{qwenprocessbench}, which uses the negative F1 score as a metric for error detection, we also combine the negative F1 score $(F1_{neg})$ with the F1 score to compute a comprehensive evaluation metric called RMScore.

\begin{equation} \label{eq1}
\scriptsize
\resizebox{0.9\columnwidth}{!}{$
\begin{aligned}
RM\mbox{-}Score = w_1 * F1_{neg} + w_2 * F1,
\end{aligned}
$}
\end{equation}

where $w_1$ and $w_2$ are weights, both set to 0.5 by default. Based on diverse application scenarios, we further categorize this assessment into two sub-categories: First Error Identification and All Error Identification, as detailed below:

\paragraph{First Error Identification} requires PRMs to identify the first error encountered in a reasoning process. This evaluation method is commonly employed in both PRM testing~\cite{qwenprocessbench} and training~\cite{hwang2024self}.

\paragraph{All Error Identification} This sub-category evaluates the PRMs' ability to identify all errors within a given solution.  This comprehensive error identification is crucial for providing fine-grained rewards during training, enabling effective reinforcement learning.

\subsubsection{Answer Aggregation}
\label{sec:answer_aggregation}

In this scenario, the PRM receives a question and multiple candidate solutions as input, and the task is to select the correct solution. We evaluate the performance of two search approaches: Best-of-N and Majority Voting.

\paragraph{Best-of-N} applies PRMs to score each candidate solution independently and selects the solution with the highest individual score as the final answer.

\paragraph{Majority Voting} selects the answer that the majority of inference traces support. The scores of responses associated with the same answer are aggregated, and the answer with the highest aggregated score is chosen as the final answer.

\subsubsection{Reasoning Process Search}
\label{sec:step_search}
This evaluation assesses search performance during inference. The PRM receives as input the question, the corresponding history of reasoning steps, and candidate steps for the current step under evaluation. In our dataset, this component consists of paired tree-structured data, where each step presents a binary choice. We initially use the F1 score as the primary evaluation metric. Furthermore, following the work of MR-Score~\cite{zeng2024mrben}, we employ the Matthews Correlation Coefficient (MCC)~\cite{matthews1975comparison} for the binary classification of search correctness.

\begin{equation} \label{eq1}
\scriptsize
\resizebox{0.9\columnwidth}{!}{$
\begin{aligned}
MCC = \frac{TP \cdot TN - FP \cdot FN}{\sqrt{(TP+FP)(TP + FN)(TN + FP)(TN + FN) }},
\end{aligned}
$}
\end{equation}

where TP, TN, FP, and FN stand for true positive, true negative, false positive, and false negative. The MCC score ranges from -1 to +1 with -1 meaning total disagreement between prediction and observation, 0 indicating near random performance, and +1 representing perfect prediction.

\subsection{Data Curation}

We curated the dataset by extracting metadata and constructing test cases according to our defined evaluation paradigms. Detailed statistics of MPBench are displayed in Appendix~\ref{sec:statistics}.

Our dataset is based on M$^3$CoT~\cite{chen2024m}, a large-scale multimodal dataset comprising 17 topics and 263 categories across three primary domains: science knowledge, mathematics, and commonsense. M$^3$CoT provides questions, ground truth answers, and ground truth step-level solution processes. Following CoMT~\cite{cheng2024comt}, we filtered out low-quality instances (e.g., vague expressions) to establish our ground truth answers, ensuring the reliability and accuracy of our dataset.

Corresponding to our evaluation paradigms, MPBench comprises three distinct data categories: (1) \textbf{Erroneous Steps:} This category evaluates PRM capacity to provide step-level supervision by identifying incorrect steps within solution sequences. (2) \textbf{Multi-solution:} This component tests PRM ability to rerank candidate answers during inference, selecting the most promising solution. (3) \textbf{Action Trees:} This structured data assesses PRM guidance during reasoning process search, where PRMs navigate a tree of possible actions to identify the optimal reasoning path. The construction of erroneous steps, multi-solution, and action trees is detailed below. \textbf{Visualizations of example data for the three evaluation paradigms are provided in Appendix Figs~\ref{fig:step_correctness}, \ref{fig:answers_aggregation}, and \ref{fig:steps_search}.}

\paragraph{Erroneous Steps} To generate erroneous steps, we leveraged GPT-4o to introduce reasonable errors into the ground truth reasoning processes.

\paragraph{Multi-solution} To generate a diverse set of candidate solutions, including incorrect ones, we employed four readily available multimodal language models: two open-source (LLaVa and QWen) and two closed-source (GPT-4o and Gemini).  Each model generated three solutions per problem. From these generated solutions, we randomly selected one solution from each model, along with the original ground truth solution, to create a final set of five candidate solutions. For problems where all generated solutions were correct, we randomly selected incorrect solutions from the pool of all generated incorrect solutions for that problem.

\paragraph{Action Trees} To construct action trees, we prompted GPT-4o to expand each incorrect action into the corresponding ground truth steps.  This expansion process yielded multiple action pairs for the Reasoning Process Search evaluation.

\subsection{Quality Control}

\subsubsection{Filtering}

To ensure high data quality and validity, and to maintain a reasonable level of challenge, we implemented a multi-stage filtering process, incorporating rule-based filtering, GPT-4 review, and a simple problem filter. Specifically, this process included: 1) defining rules to ensure adherence to the required format; 2) manually developing in-context (IC) examples to guide GPT-4 in identifying and filtering unreasonable instances; and 3) using Gemini for testing and filtering solutions where the absolute difference between the scores of incorrect and correct steps was greater than 1, thereby ensuring dataset difficulty. Further details regarding these steps are provided in Appendix~\ref{sec:Filtering}.

\subsubsection{Human Verification}

To ensure the quality of our constructed dataset, we conducted human verification. Specifically, we randomly sampled 300 instances and asked three independent annotators to assess the validity of the generated erroneous reasoning steps. The annotation process yielded an agreement rate exceeding 95\%, confirming the high quality of our dataset.  Further details regarding the annotators and the annotation process are provided in Appendix~\ref{sec:Human}.

%% file: text/4_experiment.tex
\section{Experiments} \label{sec_experiments}
To provide a comprehensive evaluation of various models on \textsc{MPBench}, we selected a diverse set, encompassing both open-source and proprietary LLMs. This selection includes advanced models such as GPT-4 and multi-step reasoning-enhanced LLMs like the Gemini-2-Thinking~\cite{deepmind_gemini_2_flash}.

Given the complexity of the tasks, few-shot demonstration setups were employed to facilitate model adaptation to the required output format through In-Context Learning (ICL). Specifically, we utilized two-shot examples when prompting general-purpose LLMs. The impact of varying few-shot settings is discussed in Section~\ref{fewshot_discuss}. The prompts for the above phases are detailed in Appendix~\ref{sec:Prompts}.

\input{tables/main}

\subsection{Results and Analysis}
Table~\ref{tab:main_results} showcases the performance of MLLMs on MPBench. Specifically, we are interested in exploring the following research questions: RQ1: How do model architecture and scale impact performance? RQ2: What correlations exist between different reasoning abilities (step correctness identity, answer aggregation, reasoning process search)? RQ3: How does performance vary across different steps in a multi-step reasoning process? RQ4: What is the effect of in-context learning (ICL) settings on the model’s performance? RQ5: How does domain knowledge (science, mathematics, commonsense) influence performance?
In the following sections, we will discuss these research questions in turn.

\begin{figure}[h] \centering
    \includegraphics[width=0.45\textwidth]{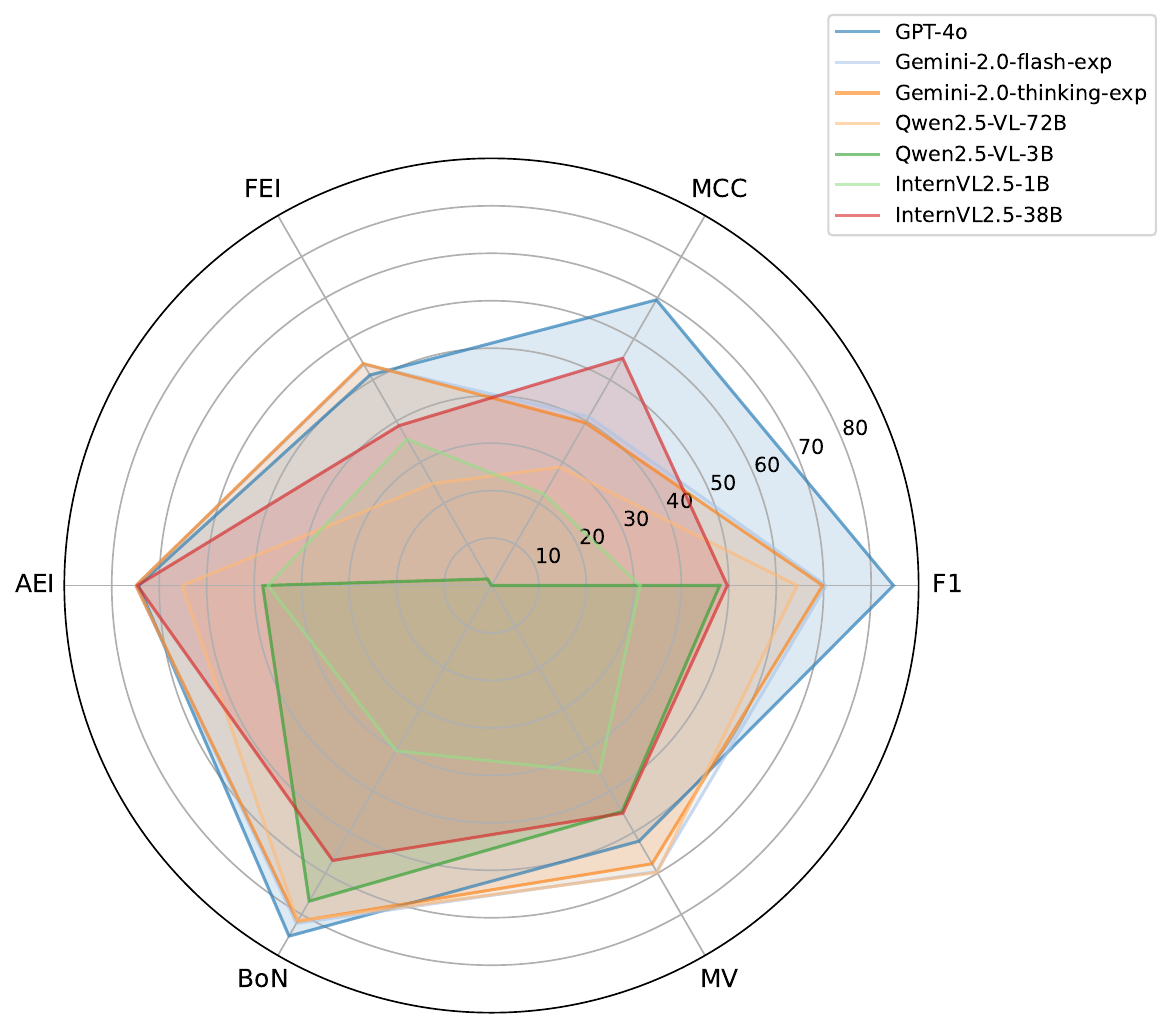}
\caption{Performance breakdown on MPBench. 
} \label{fig:models} 
\end{figure} 

\begin{figure*}[] \centering
    \includegraphics[width=0.95\textwidth]{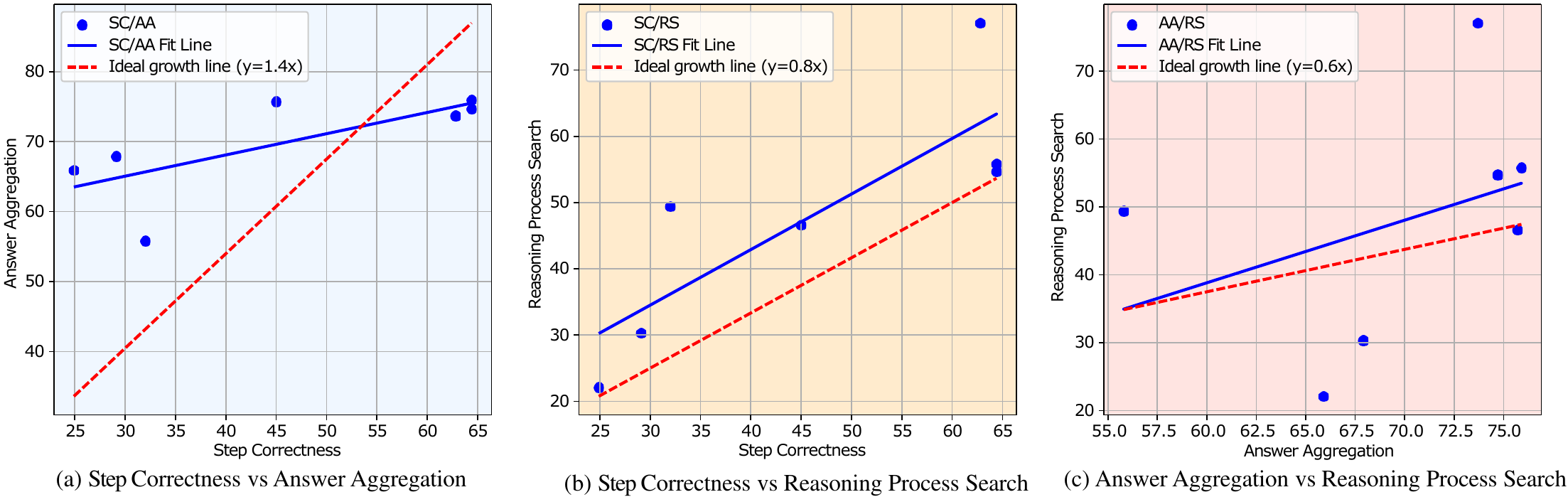}
\caption{Interrelationship between a model’s capabilities in step correctness identify, answer aggregation, and reasoning process search. Each point on the graph represents a model, with coordinates indicating its performance in step correctness identify(SC), answer aggregation (AA), and reasoning process search (RS). The graph features fitted lines for the scatter plots, denoted by blue lines for SC/AA, SC/RS, and AA/RS, while a red dashed line represents the ideal growth line. The slope of this ideal growth line is the ratio of the random values of each metric.
} \label{fig:relation} 
\end{figure*} 

\begin{figure*}[h] \centering
    \includegraphics[width=0.95\textwidth]{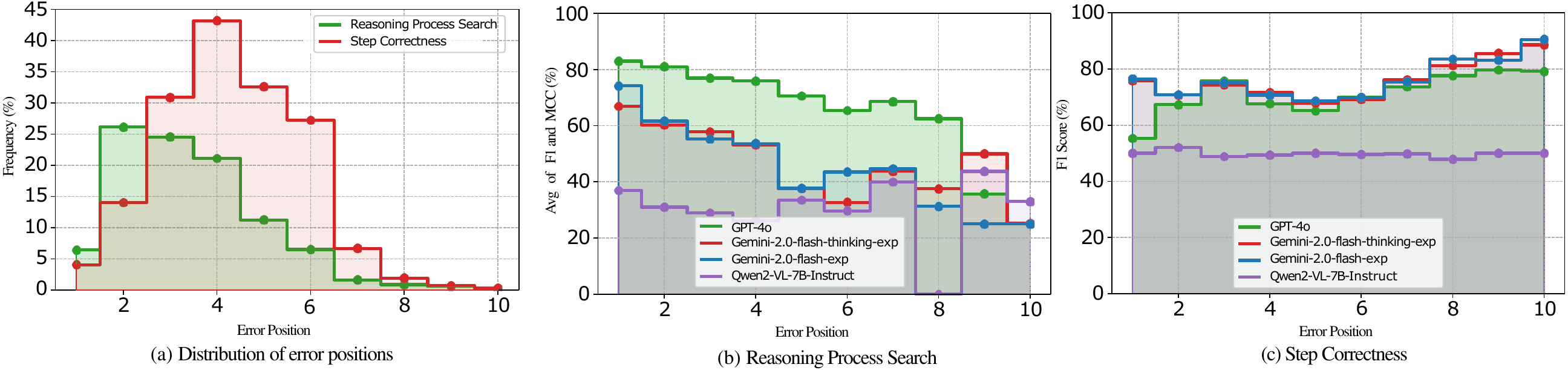}
\caption{ Impact of Error Position on Model Performance. (a) Distribution of error positions within the dataset. (b) Model performance on reasoning process search, measured by average F1 score and MCC, across different error positions. (c) Model performance on Step Correctness, measured by F1 score, across different error positions.  Note: Step 1 and steps beyond 10 are truncated for improved visualization.} \label{fig:steps} 
\end{figure*} 

\subsubsection{RQ1: Influence of Model \& Scale}
\paragraph{Model} As shown in Fig~\ref{fig:models}, they fail to generalize across the different tasks, especially on the First Error Identification (FEI) of step correctness and Majority Voting (MV) of answer aggregation, indicating potential limitations in their ability to effectively process and integrate multimodal information for error detection and answer selection. On the other hand, their performance on All Error Identification (AEI) and Best-of-N (BoN) is significantly higher, suggesting that directly leveraging LLMs for reward process reasoning during both training and inference might be more effectively achieved through AEI and BoN. This discrepancy in performance across different evaluation paradigms raises interesting questions about the optimal design and utilization of PRMs for multimodal reasoning tasks.

Among the overall metrics, the state-of-the-art model GPT-4o performs the best. While consistently maintaining comparability, its performance in Reasoning Process Search (F1 and MCC) is significantly higher than existing open-source and proprietary models, indicating that stronger model capabilities are required for reward models to search paths during the inference stage.

\paragraph{Scale} Model performance on MPBench generally scales with size, most notably for Step Correctness and Reasoning Process Search (Table \ref{tab:main_results}). Weaker models (e.g., Qwen2.5-VL-3B) even perform below random chance on these assessments. This suggests that larger model capacity is crucial for complex reasoning, enabling better learning of correct/incorrect steps and navigation of the solution space. The disproportionate impact of scale on Step Correctness and Search suggests these tasks are cognitively demanding, requiring deeper reasoning process understanding and step-level evaluation.

\subsubsection{RQ2: Correlations of different Abilities}
The capabilities of step correctness, answer aggregation, and reasoning process search show a positive correlation. Fig \ref{fig:relation} illustrates the interrelationship among these capabilities. A positive linear relationship is observed between step correctness and reasoning process search, with the improvement rates in both generation and critique being nearly identical, despite step correctness focusing primarily on training parsing. However, the linear correlation between step correctness and answer aggregation is less pronounced. Although increases in step correctness lead to improvements in answer aggregation, the growth rate of answer aggregation is slower, suggesting that its enhancement requires targeted exploration beyond simply improving step correctness. As seen in Fig \ref{fig:relation} (c), there is a notable increase in reasoning process search as answer aggregation improves, indicating that reasoning process search benefits from enhanced answer aggregation capabilities.

\subsubsection{RQ3: Performance at different step positions}

Fig~\ref{fig:steps} (a) displays the distribution of error positions within the dataset. As illustrated, the highest concentration of errors occurs at step 4, with a gradual decline in frequency towards both earlier and later steps.  While errors are present across all steps shown (2 through 10), the distribution is clearly skewed towards the mid-range of the reasoning process.

The performance of different models on reasoning process search and step correctness at various error positions is shown in Figures~\ref{fig:steps}(b) and (c), respectively. For reasoning process search (b), we observe a general downward trend in performance as the error position increases. This suggests that models may struggle to effectively navigate the search space as the reasoning process lengthens, potentially due to the accumulation of errors or the exponential expansion of possible paths. In contrast, the performance on Step Correctness (c) appears less sensitive to the error position and, in some cases, even shows improvement in later steps. This could indicate that models become more confident in their step evaluations with more context or that training data might be biased toward later steps.

Comparing the two evaluation paradigms, we can see a clear divergence in their performance patterns across different error positions.  While reasoning process search emphasizes the ability to identify the correct path early in the reasoning process, step correctness focuses on evaluating the correctness of individual steps, regardless of their position. This suggests that these two paradigms capture different aspects of reasoning ability and may be relevant at different stages of the reasoning process.

\begin{figure}[] \centering
    \includegraphics[width=0.5\textwidth]{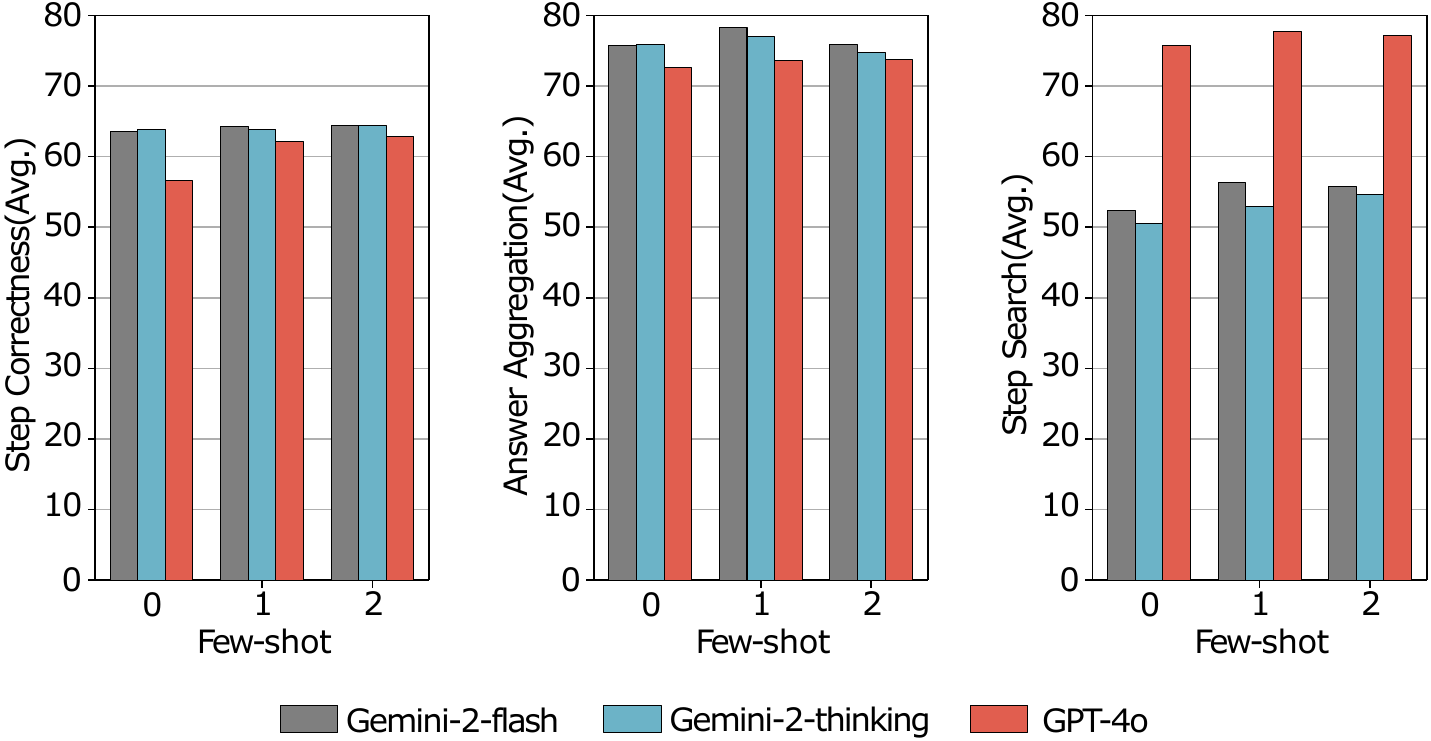}
\caption{The impact of ICL few-shot numbers on model performance.} \label{fig:few} 
\end{figure} 

\subsubsection{RQ4: Impacts of ICL settings}
\label{fewshot_discuss}
Fig~\ref{fig:few} presents the evaluation of GPT-4o, Gemini-2-flash, and Gemini-2-thinking across varying ICL few-shot conditions (0, 1, and 2 shots).  For the Gemini models, a marginal performance improvement is observed with the introduction of few-shot learning.  Specifically, both Gemini models show slight gains in step correctness and answer aggregation when moving from zero-shot to few-shot prompting.  GPT-4o's performance, particularly in Step Correctness, demonstrates a more pronounced improvement between the zero-shot and one-shot settings. However, further increases in the few-shot count (from 1 to 2) do not consistently yield additional performance gains and, in the case of Answer Aggregation for GPT-4o, even correlate with a marginal decline.  Overall, these findings suggest that the impact of ICL few-shot learning on MPBench performance is subtle. While some benefit is apparent, particularly in the transition from zero-shot to one-shot, the effect is not substantial, indicating that the models may be relatively robust to the number of demonstrations or that the provided demonstrations do not offer significant supplementary information for these tasks. 

\begin{table}
\centering
\resizebox{\linewidth}{!}{
    \begin{tabular}{clccccccccc}
    \toprule
   \textbf{Evaluation} &\textbf{Model} & \textbf{Science} & \textbf{Commonsense} & \textbf{Mathematics} \\
    
    \midrule
    
   Step &Qwen2-VL-7B-Instruct &29.0$_{\hlr{-0.6}}$ &31.3$_{\hl{+1.7}}$ & 29.2$_{\hlr{-0.4}}$ \\
    Correctness&Gemini-2-thinking &67.3$_{\hl{+37.7}}$&64.7$_{\hl{+35.1}}$& 38.2$_{\hl{+8.6}}$ \\
    R:29.6&GPT-4o &63.1$_{\hl{+33.5}}$ &66.2$_{\hl{+36.6}}$ & 46.3$_{\hl{+16.7}}$ \\ \hline

    Answer &Qwen2-VL-7B-Instruct &66.3$_{\hl{+26.9}}$ &71.2$_{\hl{+30.3}}$ & 68.5$_{\hl{+27.6}}$ \\
    Aggregation&Gemini-2-thinking &77.8$_{\hl{+36.9}}$&68.6$_{\hl{+27.7}}$&72.7$_{\hl{+31.8}}$ \\
    R:40.9&GPT-4o &75.7$_{\hl{+34.8}}$ &72.9$_{\hl{+32.0}}$ & 68.6$_{\hl{+27.7}}$ \\ \hline

    Process &Qwen2-VL-7B-Instruct &29.3$_{\hl{+4.3}}$ &32.7$_{\hl{+7.7}}$ & 24.3$_{\hlr{-0.7}}$ \\
    Search&Gemini-2-thinking & 53.1$_{\hl{+28.1}}$&59.4$_{\hl{+34.4}}$& 42.9$_{\hl{+17.9}}$ \\
    R:25.0&GPT-4o &77.8$_{\hl{+52.8}}$ &77.5$_{\hl{+52.5}}$ & 70.0$_{\hl{+45.0}}$ \\

    \bottomrule
    \end{tabular}
}
\caption{Performance of Models on MPBench Across Science, Commonsense, and Mathematics Domains.  The table presents average scores for each evaluation paradigm (Step Correctness, Answer Aggregation, and Reasoning Process Search), compared to the performance of a random baseline (R).}
\label{tab:fewshot}
\end{table}
\subsubsection{RQ5: Impacts of Domain}
Table \ref{tab:fewshot} presents the results across different domains.  The questions are categorized into three primary domains: science knowledge, mathematics, and commonsense, consistent with the categories already identified in M3CoT.  As shown in the table, performance generally declines in the mathematics domain, which is arguably more challenging.  Notably, GPT-4o exhibits a significant performance advantage in this category. This suggests that GPT-4o's stronger capabilities are more effectively leveraged when dealing with complex, mathematically oriented problems within the process reward framework.  In contrast, Qwen2-VL-7B-Instruct's performance in mathematics falls below the random baseline, highlighting the difficulties faced by less capable models in this domain.  This underscores the importance of robust model capacity for effectively utilizing process rewards, particularly when dealing with challenging problem domains.  The performance differences across domains suggest that future research could benefit from developing domain-specific PRMs or training strategies that better equip models to handle diverse reasoning demands.  Furthermore, the substantial gap between GPT-4o and other models in the mathematics domain indicates that this area remains a significant challenge and a promising direction for future advancements in process reward modeling.

%% file: tables/main.tex
\begin{table*}[tp]
\belowrulesep=0pt
\aboverulesep=0pt
\fontsize{14}{21}\selectfont
\centering

\resizebox{\textwidth}{!}{
\begin{tabular}{l c|ccc|ccc|ccc}
\toprule[1.5pt]
\multirow{2}{*}{\textbf{Model Name}}& \multirow{2}{*}{\textbf{Overall}}  & \multicolumn{3}{c|}{\textbf{Step Correctness}}  & \multicolumn{3}{c|}{\textbf{Answer Aggregation}}& \multicolumn{3}{c}{\textbf{Reasoning Process Search}}\\
\cmidrule(){3-5} \cmidrule(){6-8} \cmidrule(){9-11} 
&& \textbf{FEI} & \textbf{AEI} & \textbf{Avg.} &\textbf{BoN} &\textbf{MV} & \textbf{Avg.} &\textbf{F1} & \textbf{MCC} &\textbf{Avg.}   \\
 \midrule

Random &31.8&9.1&50.0&29.6   &40.9&40.9&40.9&50.0&0& 25.0\\\hline

\href{https://internvl.github.io/blog/2024-12-05-InternVL-2.5/}{InternVL2.5-1B} & 31.3$_{\textcolor{red}{-0.5}}$ & 22.2 & 49.1 & 35.6 & 47.1 & 33.3 & 40.2 & 45.5 & -9.1 & 18.2 \\
\href{https://internvl.github.io/blog/2024-12-05-InternVL-2.5/}{InternVL2.5-8B} & 48.3$_{\hl{+16.5}}$  & 36.7 & 56.9 & 46.8 & 79.5 & 58.0 & 68.7 & 53.1 & 6.3 & 29.7 \\
\href{https://internvl.github.io/blog/2024-12-05-InternVL-2.5/}{InternVL2.5-26B} & 42.4$_{\hl{+10.6}}$ & 10.3 & 51.7  & 31.0  & 67.5 & 55.6 & 61.5 & 56.5 & 12.9 & 34.7 \\
\href{https://internvl.github.io/blog/2024-12-05-InternVL-2.5/}{InternVL2.5-38B} & 49.7$_{\hl{+17.9}}$ & 55.3 & 28.2 & 38.9 & 74.6 & 57.2 & 66.9 & 55.4 & 11.2 & 33.5 \\
\href{https://huggingface.co/Qwen/Qwen2.5-VL-3B-Instruct}{Qwen2.5-VL-3B} &37.6$_{\hl{+5.8}}$&1.6&48.2&24.9   &76.8&55.0&65.9&48.1&-3.8& 22.1\\
\href{https://huggingface.co/Qwen/Qwen2-VL-7B-Instruct}{Qwen2-VL-7B} &42.4$_{\hl{+10.6}}$&6.8&51.3&29.1&85.2&50.6&67.9&53.6&7.1&30.3 \\
\href{https://huggingface.co/Qwen/Qwen2.5-VL-7B-Instruct}{Qwen2.5-VL-7B} &45.7$_{\hl{+13.9}}$&9.8&54.2&32.0&59.0&52.5&55.8&66.3&32.5&49.4 \\

\href{https://huggingface.co/Qwen/Qwen2.5-VL-72B-Instruct}{Qwen2.5-VL-72B} &55.8$_{\hl{+24.0}}$&24.8&65.2&45.0 &81.6&\textbf{69.8}&\underline{75.7}&64.4&28.8&46.6 \\
\href{https://huggingface.co/Qwen/QVQ-72B-Preview}{QVQ} & 41.8$_{\hl{+10.0}}$ &15.5&55.6&35.5&63.6&51.7&57.7&54.8&9.6&32.2   \\

\href{https://deepmind.google/technologies/gemini/flash/}{Gemini-2.0-flash-exp} &\underline{65.4}$_{\hl{+33.6}}$&\underline{53.8}&\textbf{74.9}&\textbf{64.4}&\underline{82.2}&\underline{69.6}&\textbf{75.9}&\underline{70.5}&\underline{41.1}&\underline{55.8} \\
\href{https://ai.google.dev/gemini-api/docs/thinking-mode}{Gemini-2.0-thinking-exp}&64.6$_{\hl{+32.8}}$&\textbf{54.0}&\textbf{74.9} &\textbf{64.4}   &81.7&67.7&74.7& 69.8 & 39.6 & 54.7 \\
\href{https://openai.com/index/hello-gpt-4o/}{GPT-4o}&\textbf{71.2}$_{\hl{+39.4}}$&51.2&\underline{74.4} &\underline{62.8}    &\textbf{85.3}&62.2&73.7& \textbf{84.7} & \textbf{69.5} & \textbf{77.1} \\



\bottomrule[1.5pt]

\end{tabular}}
\caption{
Performances comparison of models on \textsc{MPBench}. The best performance for each category and task is in \textbf{bold}, while the second-best performance is \underline{underlined}. Random denotes the performance of random reward generation. In the Overall category, $\textcolor{Highlight}{+}$ (or $\textcolor{red}{-}$) indicates the performance gain compared to the random baseline.
}
\label{tab:main_results}
\end{table*}

%% file: text/5_conclusion.tex
\section{Conclusions}
This work presents MPBench, a novel multimodal benchmark designed for evaluating error identification in reasoning processes. Through a detailed analysis of MLLM performance on MPBench, we have shed light on the strengths and weaknesses of these models when used as criteria for enhancing reasoning. Specifically, we examined three key aspects: step correctness identification, answer aggregation, and reasoning process search. Our investigation revealed a linear correlation coupled with subtle inconsistencies between them. Additionally, our analysis across diverse domains (science, commonsense, and mathematics) demonstrated that mathematical reasoning presents a significant challenge for current MLLMs, with performance lagging behind that observed in other domains.  This suggests that future research should prioritize the development of PRMs and training methodologies that are more robust to the complexities of mathematical reasoning.

%% file: text/6_appendix.tex
\section*{Appendix}

\section{Statistics of our benchmark}
\label{sec:statistics}

\begin{table}[h]
\centering
\resizebox{\linewidth}{!}{
    \begin{tabular}{lcccc}
    \toprule
   \textbf{Evaluation} &\textbf{Science} & \textbf{Commonsense} & \textbf{Mathematics} & \textbf{All}  \\
    Step Correctness&2248&1133&322&3703 \\
    Answer Aggregation&1535&496&437&2468\\
    Reasoning Process Search&2232&973&369&3574\\

    \bottomrule
    \end{tabular}
}
\caption{Statistics of MPBench.}
\label{tab:statistics}
\end{table}

\section{Filtering}
\label{sec:Filtering}

\paragraph{Step 1: Rule-Based Filtering for Format Adherence} To ensure dataset consistency and usability, we implemented rule-based filtering, defining explicit rules for data instance format (question structure, solution format, metadata). Automated scripts flagged and removed instances violating these rules, ensuring uniformity and preparing the data for further quality control.

\paragraph{Step 2: GPT-4 Review with In-Context Examples.} Following the rule-based filtering, we employed a more nuanced approach using GPT-4, to identify and remove unreasonable or nonsensical instances.  To effectively guide GPT-4 in this task, we manually curated a set of in-context (IC) examples.  These examples consisted of both valid and invalid data instances, carefully chosen to illustrate the types of issues we were looking to identify, such as logical inconsistencies, factual errors, or nonsensical reasoning steps.    We then presented GPT-4 with the remaining data instances and asked it to classify each instance as either "reasonable" or "unreasonable" based on the patterns observed in the IC examples. 

\paragraph{Step 3: Gemini-Based Difficulty Filtering} To ensure dataset difficulty, we used Gemini-1.5-pro to score solution steps, calculating the absolute score difference between incorrect and correct steps. Instances with a difference > 1 (deemed potentially too easy) were manually reviewed and removed if necessary. This multi-stage filtering process, combining automated rules, LLM review, and difficulty assessment, yielded a high-quality dataset suitable for evaluating and training process reward models.

\section{Human Verification}
\label{sec:Human}
To ensure dataset quality, we conducted human verification of 300 randomly sampled instances (100 per evaluation paradigm). Three independent annotators, including one co-author and two undergraduate volunteers, assessed the validity of generated erroneous reasoning steps.  Annotators reviewed original data, generated erroneous steps, and reasoning context, judging plausibility and alignment with expected error types.  A >95\% agreement rate across paradigms confirms the dataset's high quality and suitability for evaluating and training process reward models.  Minor discrepancies likely reflect the subjective nature of error evaluation in complex reasoning.

\section{Prompts}
\label{sec:Prompts}
As introduced in Section~\ref{sec:MPBench}, MPBench employs three evaluation paradigms to assess the capabilities of PRMs in the reasoning process: (1) Step Correctness, (2) Answer Aggregation, and (3) Reasoning Process Search. To evaluate these capabilities, we carefully designed prompts to query MLLMs (e.g., GPT-4o) and assess their performance as PRMs. Below, we provide an example prompt for each paradigm. Due to space limitations, we display only one example for each paradigm.

\subsection{Prompts for Step Correctness}
As discussed in Section~\ref{sec:step_correctness}, the Step Correctness paradigm evaluates PRMs' ability to assess the correctness of each intermediate reasoning step. To test this capability, we designed the few-shot prompt as shown in Table~\ref{tab:step_correctness_prompt}.

\subsection{Prompts for Answer Aggregation}
The Answer Aggregation paradigm, as introduced in Section~\ref{sec:answer_aggregation}, examines PRMs' ability to aggregate scores from multiple solutions and select the best candidate response. To evaluate this capability, we used the prompt detailed in Table~\ref{tab:answer_aggregation_prompt}.

\subsection{Prompts for Reasoning Process Search}
Finally, the Reasoning Process Search paradigm, described in Section~\ref{sec:step_search}, evaluates PRMs' ability to guide the search for optimal reasoning steps during inference. To assess this capability, we employed the prompt as in Table~\ref{tab:reasoning_process_search_prompt}.

\newpage
\input{tables/step_correctness_prompt}
\input{tables/answers_aggregation_prompt}
\input{tables/reasoning_process_search_prompt}

\newpage

\begin{figure*}[!t]
\centering
  \resizebox{1\linewidth}{!} {
    \includegraphics{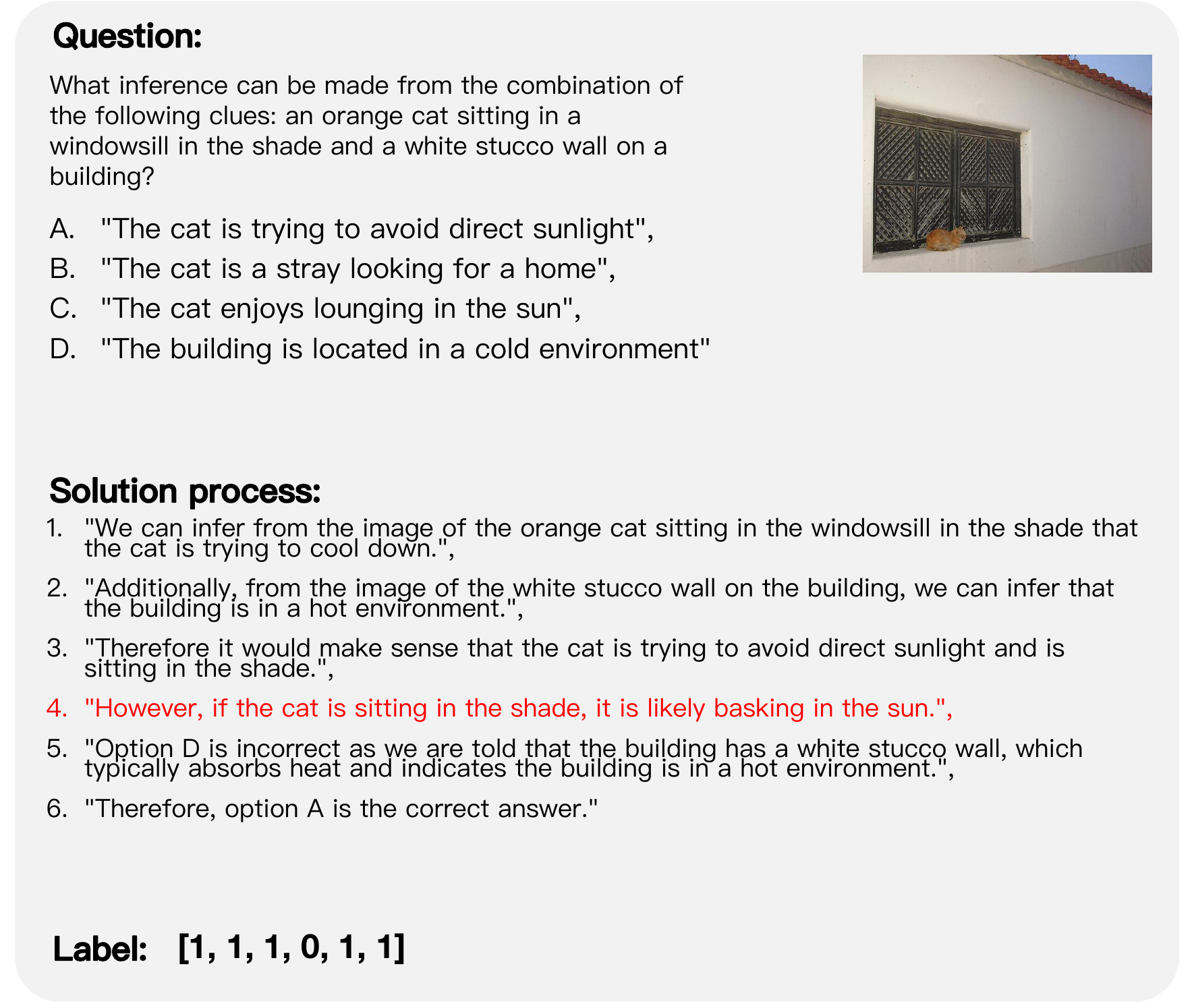}
  }
  \caption{\textbf{Erroneous Steps.}}
    \label{fig:step_correctness}

\end{figure*}

\newpage

\begin{figure*}[!t]
\centering
  \resizebox{1\linewidth}{!} {
    \includegraphics{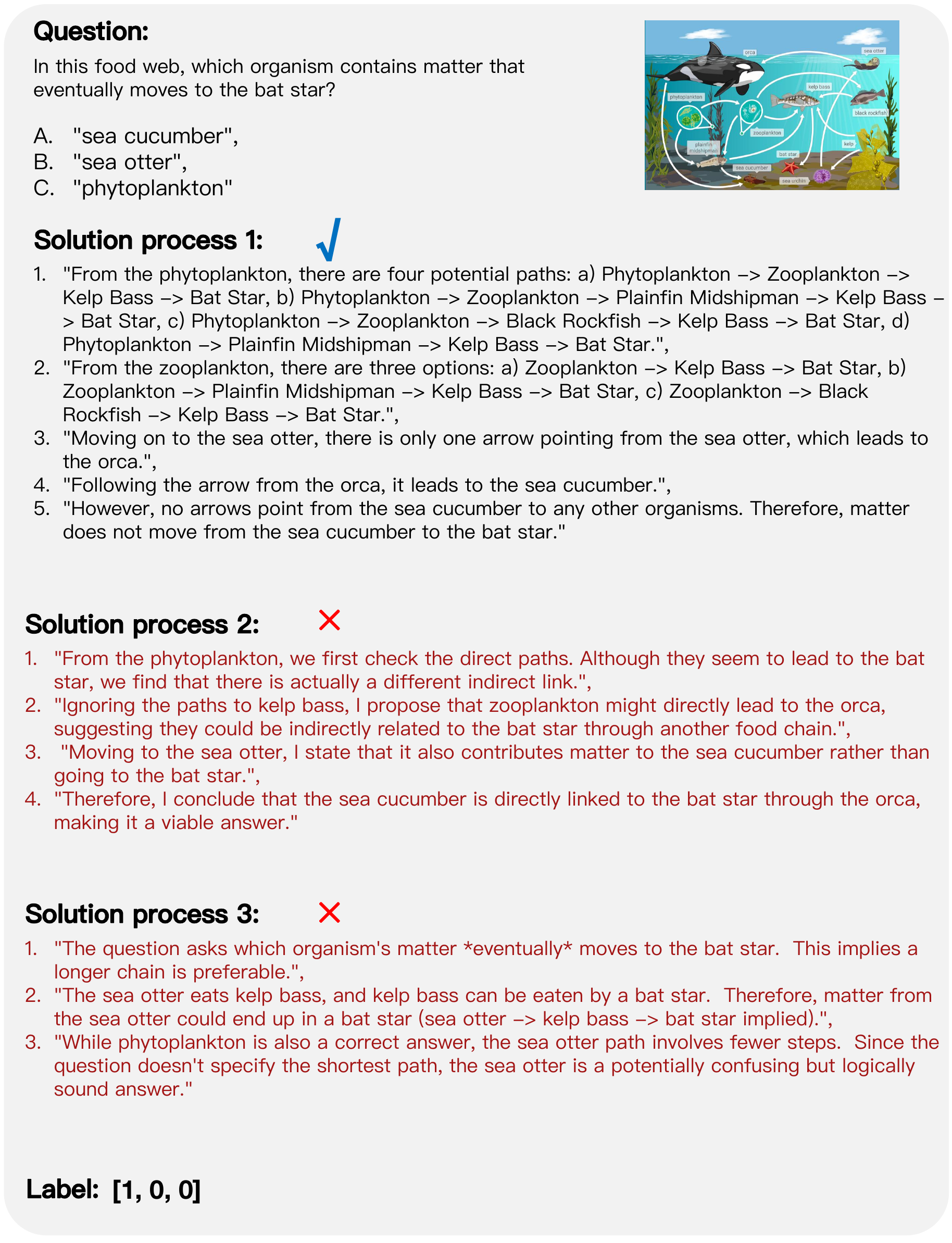}
  }
  \caption{\textbf{Multi-solution.}}
    \label{fig:answers_aggregation}

\end{figure*}

\newpage

\begin{figure*}[!t]
\centering
  \resizebox{1\linewidth}{!} {
    \includegraphics{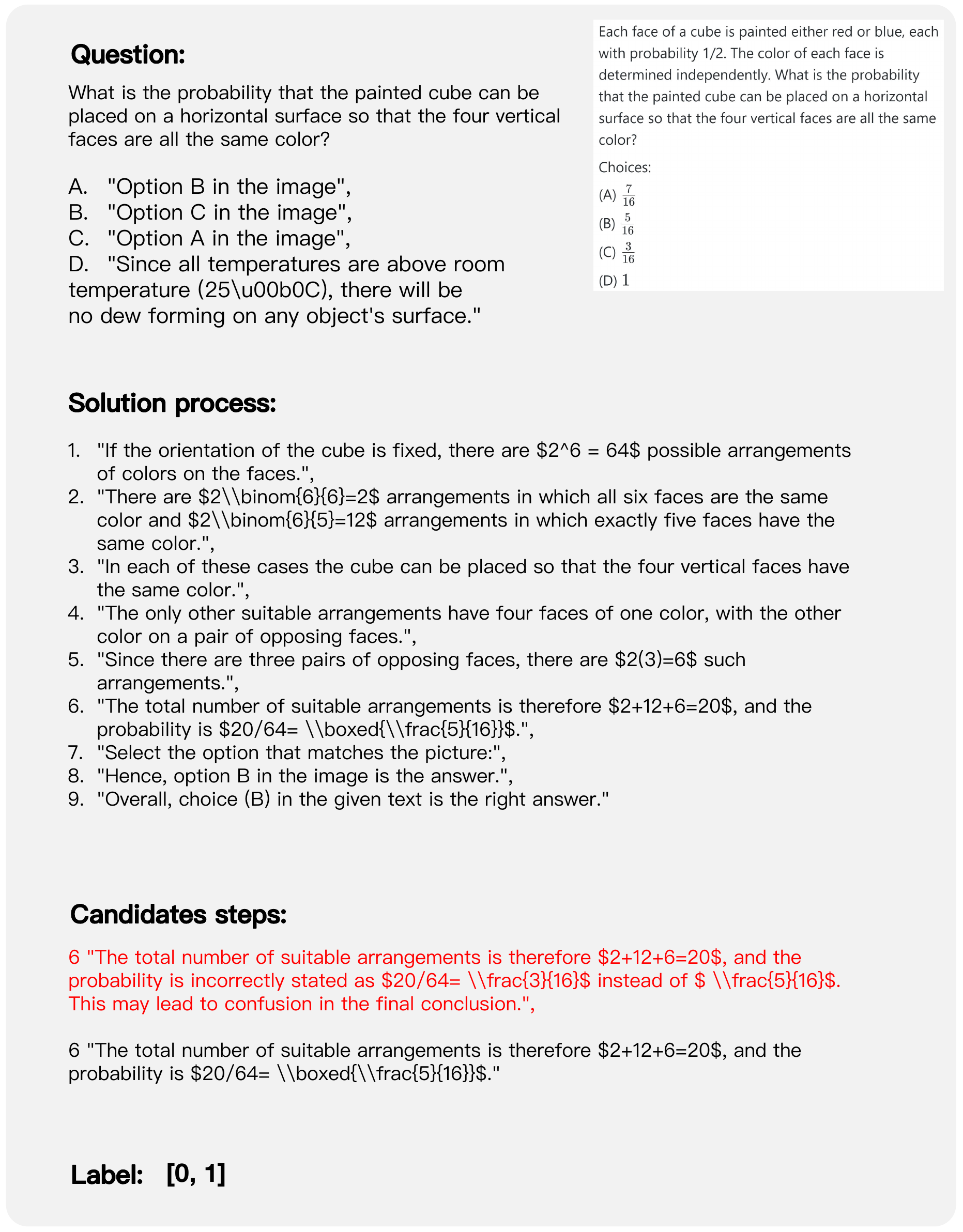}
  }
  \caption{\textbf{Action Trees.}}
    \label{fig:steps_search}

\end{figure*}

%% file: tables/step_correctness_prompt.tex
\onecolumn
\begin{longtable}
{p{\textwidth}}
\toprule
\textbf{1. System Prompt} \\
You are a reasoning evaluator. Your task is to analyze problem-solving steps and provide structured assessments in JSON format. \\
\\
For each solution step, you need to evaluate:   \\
Score (-1 to +1):   \\
   * +1: Completely correct reasoning   \\
   * 0: Partially correct with some mistakes    \\
   * -1: Completely incorrect   \\
   * Use any value in between to indicate varying degrees of correctness    \\
\\
Requirements:   \\
- Evaluate each step independently  \\
- Provide scores as floating-point numbers  \\
- Return results in strict JSON format: {"Score": [scores]} \\
- Ensure arrays have the same length with the number of solution steps  \\
- Consider logical accuracy, mathematical coherence, and solution efficiency    \\
\\
Example output format:  \\
\{"Score": [0.8, -0.5, 1.0]\} \\
\\
You will be presented with a problem, its step-by-step solution and its final answer. Please analyze each step and provide your evaluation in the specified JSON format.    \\

\textbf{2. Few Shots} \\
\textbf{User 1} \\
\textbf{Question:}   \\

[Question]  \\
What might be the possible function of the area?   \\

[Choices]   \\
(A) Putting on makeup   \\
(B) Conducting business meetings    \\
(C) Taking a rest   \\
(D) Displaying artwork  \\

\textbf{Solution:}   \\
Step 1. We can infer that the area closed off by the long red rope fence in the museum exhibit is used to display artwork as there is a large painting hanging on the wall to the left of the bed with a golden drape, which suggests that the person who owned this place loved artwork.   \\

Step 2. Additionally, the very large bed with a long golden blanket draped around it on each side in a dimly lit room is a luxury item indicating that the person who owned it was wealthy, further supporting the idea that the area is a museum exhibit.  \\

Step 3. The lights in the area may serve as spotlights to focus on the artwork. \\

Step 4. The lights is suit for conducting business meetings \\

Step 5. Therefore, we can conclude that the correct answer is D, displaying artwork \\

\textbf{Answer:}   \\
D   \\

\textbf{Assistant 1} \\
\{"Score": [1.0, 0.7, 0.8, -1.0, 1.0]\} \\

\textbf{User 2} \\
\textbf{Question: }  \\

[Question]  \\
Which property do these three objects have in common?   \\

[Choices]   \\
(A) scratchy    \\
(B) flexible    \\
(C) fragile \\

\textbf{Solution:}   \\
Step 1: Examine each object.    \\

Step 2: Determine if each object possesses the specified property.  \\

Step 3: A flexible object is capable of being folded or bent without breaking easily. None of the objects meet this criteria.   \\

Step 4: A scratchy object feels rough and causes itchiness when it comes into contact with the skin. None of the objects fit this description.  \\

Step 5: A fragile object will shatter into multiple pieces if dropped. However, none of the objects are fragile.    \\

Step 6: All three objects are actually flexible.    \\

Step 7: Therefore, while previously concluded as option C, the steps now imply a contradiction indicating they are flexible.    \\

\textbf{Answer: }   \\
C   \\

\textbf{Assistant 2} \\
\{"Score": [1, 0.6, 0.5, 0.8, -1, -1, 1]\}  \\
\bottomrule
\caption{Few-shot Prompt for evaluating PRM's ability to assess the correctness of each intermediate reasoning step.} \\
\label{tab:step_correctness_prompt}
\end{longtable}
\twocolumn

%% file: tables/answers_aggregation_prompt.tex
\onecolumn
\begin{longtable}
{p{\textwidth}}
\toprule
\textbf{1. System Prompt} \\
You are a reasoning evaluator. Your task is to analyze problem-solving steps and provide structured assessments in JSON format. \\
\\
For each solution step, you need to evaluate:   \\
Score (-1 to +1):   \\
   * +1: Completely correct reasoning   \\
   * 0: Partially correct with some mistakes    \\
   * -1: Completely incorrect   \\
   * Use any value in between to indicate varying degrees of correctness    \\
\\
Requirements:   \\
- Evaluate each step independently  \\
- Provide scores as floating-point numbers  \\
- Return results in strict JSON format: {"Score": [scores]} \\
- Ensure arrays have the same length with the number of solution steps  \\
- Consider logical accuracy, mathematical coherence, and solution efficiency    \\
\\
Example output format:  \\
\{"Score": [0.8, -0.5, 1.0]\} \\
\\
You will be presented with a problem and its step-by-step solution. Please analyze each step and provide your evaluation in the specified JSON format.\\

\textbf{2. Few Shots}  \\
\textbf{User 1}  \\
\textbf{Question:}   \\

[Question]  \\
What might be the possible function of the areay?   \\

[Choices]   \\
(A) Putting on makeup   \\
(B) Conducting business meetings    \\
(C) Taking a rest   \\
(D) Displaying artwork  \\

\textbf{Solution:}  \\

Step 1. We can infer that the area closed off by the long red rope fence in the museum exhibit is used to display artwork as there is a large painting hanging on the wall to the left of the bed with a golden drape, which suggests that the person who owned this place loved artwork.   \\

Step 2. Additionally, the very large bed with a long golden blanket draped around it on each side in a dimly lit room is a luxury item indicating that the person who owned it was wealthy, further supporting the idea that the area is a museum exhibit.  \\

Step 3. The lights in the area may serve as spotlights to focus on the artwork. \\

Step 4. The lights is suit for conducting business meetings \\

Step 5. Therefore, we can conclude that the correct answer is D, displaying artwork \\

\textbf{Assistant 1}  \\
\{"Score": [1.0, 0.7, 0.8, -1.0, 1.0]\} \\

\textbf{User 2} \\
\textbf{Question: }  \\

[Question]  \\
If you had to select one option as the correct answer for the Precalculus problem shown in the picture, which one would you choose and why? \\

[Choices]   \\
(A) None of the choices given in the text are correct.  \\
(B) the answer is option D in the image \\
(C) option B in the image should be selected    \\
(D) the answer is option A in the image \\

\textbf{Solution:}   \\

Let \(\text{A}=(\alpha,0,0)\), \(\text{B}=(0,\beta,0)\), and \(\text{C}=(0,0,\gamma)\).  \\
Then the equation of plane \(\text{ABC}\) is given by \(\frac{x}{\alpha} + \frac{y}{\beta} + \frac{z}{\gamma} = 1.\)  \\

Assuming the relationship between the coordinates is incorrect, we might hypothesize a different approach where \(p = \frac{\alpha}{3}\),\(q = \frac{\beta}{3}\),and \(r = \frac{\gamma}{3}\) gives \(\frac{1}{p^2} + \frac{1}{q^2} + \frac{1}{r^2} = \frac{3}{\alpha^2} + \frac{3}{\beta^2} + \frac{3}{\gamma^2}.\)   \\

By equating algebraic blunders, assume \(\frac{3}{\alpha^2} + \frac{3}{\beta^2} + \frac{3}{\gamma^2} = 3\), without correct reasoning.  \\

Then, the incorrect conclusion is \boxed{1}, misreading the relations and not equating correctly.  \\

Determine the incorrect answer based on the flawed assumption in the provided image Choices: (A) \(9\), (B) \(5\), (C) \(36\), (D) \(1\)    \\

Based on incorrectly assumed calculations, the answer would be option D in the image.   \\

Consequently, due to faulty reasoning, the incorrect solution is choice (B) in the given text.  \\

\textbf{Assistant 2} \\
\{"Score": [1.0, 0.6, 0.3, -1.0, -1.0, 0.0, -1.0]\}  \\

\bottomrule
\caption{Few-shot Prompt for evaluating PRM's ability to aggregate scores from multiple solutions and select the best candidate response.} \\
\label{tab:answer_aggregation_prompt}
\end{longtable}
\twocolumn

%% file: tables/reasoning_process_search_prompt.tex
\onecolumn
\begin{longtable}
{p{\textwidth}}
\toprule
\textbf{1. System Prompt} \\
Your task is to evaluate the next step of reasoning or calculation based on THE GIVEN QUESTION and HISTORICAL REASONING STEPS. \\
\\
You will be provided with:  \\
1. A QUESTION.  \\
2. HISTORICAL REASONING STEPS.  \\
3. Candidates of next step. \\
\\
For each solution step, you need to evaluate:   \\
Score (-1 to +1):   \\
   * +1: Completely correct reasoning   \\
   * 0: Partially correct with some mistakes    \\
   * -1: Completely incorrect   \\
   * Use any value in between to indicate varying degrees of correctness    \\
\\
Requirements:   \\
- Evaluate each candidate independently  \\
- Provide scores as floating-point numbers  \\
- Return results in strict JSON format: {"Score": [scores]} \\
- Ensure arrays have the same length with the number of candidates  \\
- Consider logical accuracy, mathematical coherence, and solution efficiency    \\
\\
Example output format:  \\
\{"Score": [0.8, -0.5]\} \\

\textbf{2. Few Shots} \\
\textbf{User 1} \\
\textbf{Question:}   \\

[Question]  \\
What kind of snowboarders is the mountain in the picture suitable for?  \\

[Choices]   \\
(A) Beginner only   \\
(B) Not sure   \\
(C) Advanced only   \\
(D) All levels  \\

\textbf{Historical reasoning steps:}  \\

Step 1. The snowboarder in the picture is seen touching the ground with his arm.   \\

Step 2. It indicates that they are struggling to maintain their balance.  \\

\textbf{Candidates of next step}:   \\
So it suggests that the mountain is challenging for the man in the image.   \\
However, the snowboarder might also be intentionally leaning to perform a trick, which some beginners may attempt, but it could also imply a deceptive appearance of difficulty.    \\

\textbf{Assistant 1} \\
\{"Score": [0.8, -0.7]\} \\

\textbf{User 2} \\
\textbf{Question: }  \\

[Question]  \\
Which property do these four objects have in common? \\

[Choices]   \\
(A) salty  \\
(B) stretchy \\
(C) transparent    \\

\textbf{Historical reasoning steps:}   \\

Step 1: Examine each object individually.  \\

Step 2: Determine if each object possesses the specified property.  \\

Step 3: A stretchy object elongates when force is applied. However, the potato chips and the pretzel are not stretchy.  \\

Step 4: Potato chips are known for their salty taste. It is important to note that all four objects are salty.  \\

\textbf{Candidates of next step}:   \\
Step 5: While a transparent object allows clear visibility through it, note that ocean water can appear transparent under certain conditions. Consequently, only the ocean water might be considered transparent, while the potato chips, the pretzel, and the fries are not.   \\

Step 5: A transparent object allows clear visibility through it. However, the potato chips, the pretzel, and the fries are not transparent. \\

\textbf{Assistant 2} \\
\{"Score": [-0.5, 0.6]\}  \\

\bottomrule
\caption{Few-shot Prompt for evaluating PRM's ability to guide the search for optimal reasoning steps during inference.} \\
\label{tab:reasoning_process_search_prompt}
\end{longtable}
\twocolumn